\documentclass{article}

\PassOptionsToPackage{numbers, compress}{natbib}

\usepackage[preprint]{neurips_2025}

\usepackage[utf8]{inputenc} 
\usepackage[T1]{fontenc}    
\usepackage{hyperref}       
\usepackage{url}            
\usepackage{booktabs}       
\usepackage{amsfonts}       
\usepackage{nicefrac}       
\usepackage{microtype}      
\usepackage{xcolor}         
\usepackage{graphicx}       
\usepackage{multirow}
\usepackage{overpic}
\usepackage{tikz}
\usepackage{array}  
\usepackage{subcaption}
\title{JointSplat: Probabilistic Joint Flow-Depth Optimization for Sparse-View Gaussian Splatting}

\author{
  Yang Xiao,
  Guoan Xu, Qiang Wu, Wenjing Jia \\
  Faculty of Engineering and Information Technology\\
  University of Technology Sydney, Australia \\
  \texttt{yang.xiao-2@student.uts.edu.au}
}

\begin{document}

\maketitle

\begin{abstract}
Reconstructing 3D scenes from sparse viewpoints is a long-standing challenge 
with wide applications. 
Recent advances in feed-forward 3D Gaussian sparse-view reconstruction methods 
provide an efficient solution for real-time novel view synthesis 
by leveraging geometric priors learned from large-scale multi-view datasets and computing 3D Gaussian centers via back-projection. 
Despite offering strong geometric cues, both feed-forward multi-view depth estimation and flow-depth joint estimation face key limitations: 
the former suffers from mislocation and artifact issues in low-texture or repetitive regions, 
while the latter is prone to local noise and global inconsistency due to unreliable matches 
when ground-truth flow supervision is unavailable. 
To overcome this, we propose JointSplat, a unified framework that leverages the complementarity between optical flow and depth via a novel probabilistic optimization mechanism. Specifically, this pixel-level mechanism scales the information fusion between depth and flow based on the matching probability of optical flow during training.
Building upon the above mechanism, we further propose a novel multi-view depth-consistency loss to leverage the reliability of supervision
while suppressing misleading gradients in uncertain areas.
Evaluated on RealEstate10K and ACID, JointSplat consistently outperforms state-of-the-art (SOTA) methods, demonstrating the effectiveness and robustness of our proposed 
probabilistic joint flow-depth optimization approach for high-fidelity sparse-view 3D reconstruction. 

\end{abstract}

\section{Introduction}
\label{Introduction}

Reconstructing 3D scenes from sparse viewpoints is a long-standing challenge in computer vision, with broad applications in augmented reality, robotic navigation, and autonomous driving~\cite{huang2023visual,chen2025activegamer,lu2025gaussianlss}. Recently, 3D Gaussian Splatting (3DGS)~\cite{kerbl20233d} has emerged as an efficient scene representation technique. It models a scene as a set of sparsely distributed 3D Gaussian primitives, enabling real-time rendering from novel viewpoints. To reduce computational overhead, some methods~\cite{charatan2024pixelsplat,chen2024mvsplat,xu2024depthsplat} estimate depth via multi-view matching—leveraging geometric priors learned from large-scale multi-view datasets—and obtain 3D Gaussian centers via back-projection, enabling real-time 3D reconstruction in a single feed-forward pass. 
While depth estimated from multi-view pixel matching offers strong global structure and priors, it often struggles in scenarios with low-texture or repetitive regions~\cite{zhang2025transplat}, and lacks consistency across sparse viewpoints. In particular, on large uniform surfaces, 
depth estimation tends to diverge due to the absence of reliable texture cues~\cite{zhang2024dcpi}. 

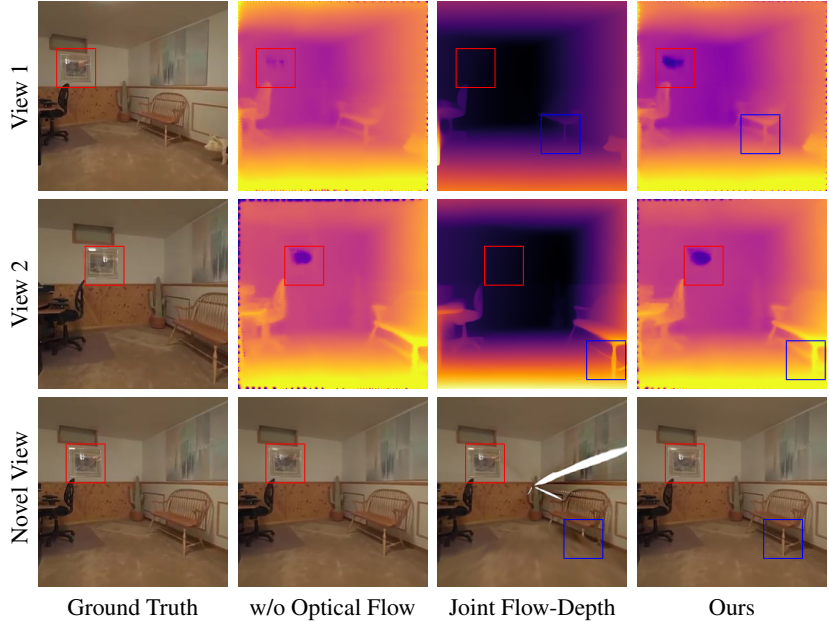
\begin{figure}[t]
  \centering
  \setlength{\tabcolsep}{2pt}   
  \renewcommand{\arraystretch}{1.0} 
  \begin{tabular}{@{}lcccc@{}}
    \rotatebox{90}{\hspace*{23pt}{\small View 1}} &
    \begin{overpic}[width=.18\linewidth]{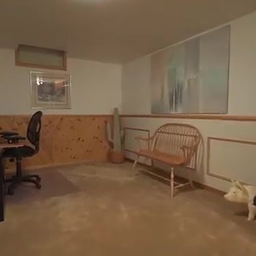}
      \put(10,55){\color{red}\framebox(20,20){}}
    \end{overpic} &
    \begin{overpic}[width=.18\linewidth]{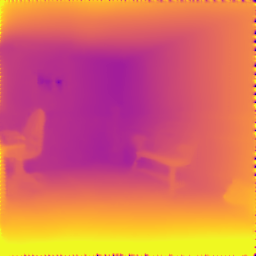}
      \put(10,55){\color{red}\framebox(20,20){}}
    \end{overpic} &
    \begin{overpic}[width=.18\linewidth]{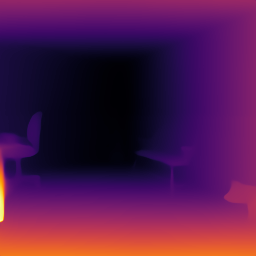}
      \put(10,55){\color{red}\framebox(20,20){}}
      \put(55,20){\color{blue}\framebox(20,20){}}
    \end{overpic} &
    \begin{overpic}[width=.18\linewidth]{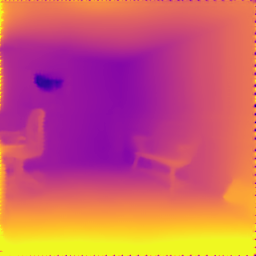}
      \put(10,55){\color{red}\framebox(20,20){}}
        \put(55,20){\color{blue}\framebox(20,20){}}
    \end{overpic} \\
    \rotatebox{90}{\hspace*{23pt}{\small View 2}} &
    \begin{overpic}[width=.18\linewidth]{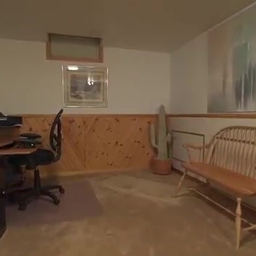}
      \put(25,55){\color{red}\framebox(20,20){}}
    \end{overpic} &
    \begin{overpic}[width=.18\linewidth]{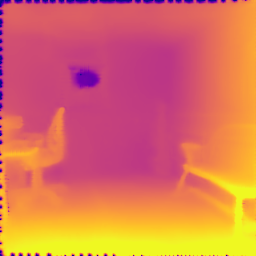}
      \put(25,55){\color{red}\framebox(20,20){}}
    \end{overpic} &
    \begin{overpic}[width=.18\linewidth]{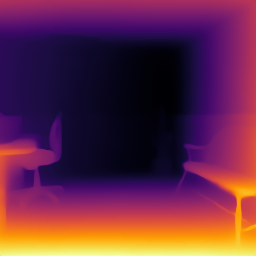}
      \put(25,55){\color{red}\framebox(20,20){}}
        \put(79,5){\color{blue}\framebox(20,20){}}
    \end{overpic} &
    \begin{overpic}[width=.18\linewidth]{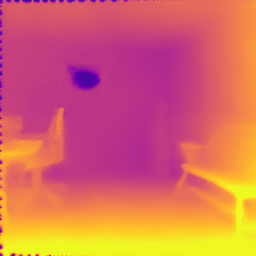}
      \put(25,55){\color{red}\framebox(20,20){}}
      \put(79,5){\color{blue}\framebox(20,20){}}
    \end{overpic} \\
    \rotatebox{90}{\hspace*{16pt}{\small Novel View}}&
    \begin{overpic}[width=.18\linewidth]{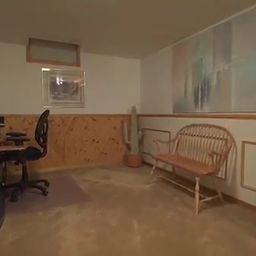}
      \put(15,55){\color{red}\framebox(20,20){}}
    \end{overpic} &
    \begin{overpic}[width=.18\linewidth]{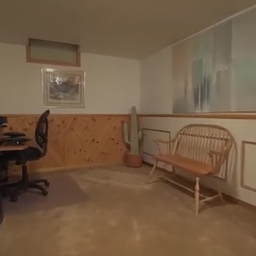}
      \put(15,55){\color{red}\framebox(20,20){}}
    \end{overpic} &
    \begin{overpic}[width=.18\linewidth]{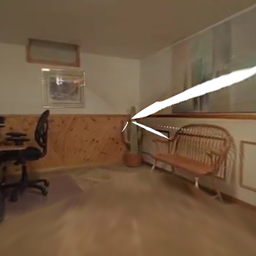}
      \put(15,55){\color{red}\framebox(20,20){}}
            \put(67,15){\color{blue}\framebox(20,20){}}
    \end{overpic} &
    \begin{overpic}[width=.18\linewidth]{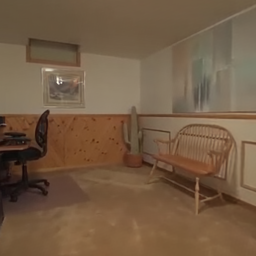}
      \put(15,55){\color{red}\framebox(20,20){}}
      \put(67,15){\color{blue}\framebox(20,20){}}
    \end{overpic} \\
    {} & 
     \small Ground Truth & \small w/o Optical Flow & \small Joint Flow-Depth & \small Ours
  \end{tabular}
  \caption{\textbf{Comparison between JointSplat and previous methods.} 
  JointSplat outperforms prior feed-forward method~\cite{xu2024depthsplat} and joint flow-depth baseline~\cite{paliwal2024coherentgs} in both depth estimation and sparse-view synthesis, especially in regions with repetitive or weak textures (see the red box), and large displacements or occlusions (see the blue box). By introducing a probabilistic optimization mechanism, our method enables more robust geometric alignment and higher-quality novel view rendering.
  }
  \label{fig:1}
  \vspace{-1.5em}
\end{figure}

In contrast, optical flow estimation typically relies on motion between temporal or stereo frames and generally incorporates global correlation and smoothness constraints. Even in regions with limited texture, it can infer pixel correspondences by enforcing local motion consistency within neighborhoods~\cite{guo2024invariant}. 
Consequently, several works have explored explicitly incorporating optical flow into 3D reconstruction pipelines~\cite{paliwal2024coherentgs,wang2020unsupervised}, leveraging its dense pixel correspondences across frames to introduce richer constraints for geometric modeling, helping mitigate the limitations of depth-only methods.
However, in the absence of ground truth flow supervision, directly incorporating optical flow may exacerbate errors: inaccurate correspondences can result in mislocation and artifacts, manifesting as local noise and global inconsistencies in the reconstructed geometry~\cite{guizilini2022learning}. 
Fig.~\ref{fig:1} illustrates this issue. 
This stems from the lack of a robust 
mechanism to model the reliability of optical flow. As such, unreliable estimations—caused by dynamic objects, occlusions, or ambiguous textures—should be identified and downscaled to avoid contaminating the depth learning signal.

To address this, some optical flow methods~\cite{sundaram2010dense,mac2012learning} rely on forward-backward consistency or predicted uncertainty to identify occluded regions, which are then excluded from loss computation or fusion. 
However, 
these approaches require separate training stages~\cite{mac2012learning}, or are unable to 
mitigate unreliable flow predictions~\cite{sundaram2010dense}.
In this paper, we propose JointSplat, a unified framework for joint flow-depth estimation aimed at high-fidelity 3D reconstruction from sparse views. Our key insight is 
a pixel-wise probabilistic optimization mechanism that dynamically scales the influence of unreliable regions (e.g., occlusions or dynamic areas) during flow estimation on depth prediction. 
This mechanism fully leverages the complementarity between optical flow and depth: 
optical flow provides pixel-level correspondence to improve local detail recovery and multi-view consistency in depth prediction, while depth estimation compensates for optical flow's structural limitations, particularly distortion in occluded regions. 
%
Our contributions are summarized as follows:
\begin{itemize}
    \item \vspace*{-0.1em}We propose a novel joint flow-depth complementary framework for high-fidelity sparse-view 3D reconstruction under unsupervised geometric constraints.
    \item \vspace*{-0.1em}We develop a probabilistic optimization mechanism to dynamically scale the influence of optical flow and depth information 
    based on optical flow's matching probability, enabling the network to favor reliable geometric cues and suppress misleading signals during training. 
    \item \vspace*{-0.1em}We propose a novel multi-view depth consistency loss, where gradient contributions are dynamically scaled by the posterior probability of cross-view image correspondences—amplifying supervision in regions with confident 
    matches while suppressing unreliable or occluded areas.
\end{itemize}

To demonstrate the efficacy of the proposed JointSplat, comprehensive experiments are conducted on public benchmarks—RealEstate10K~\cite{zhou2018stereo} and ACID~\cite{liu2021infinite}. 
On RealEstate10K~\cite{zhou2018stereo}, JointSplat achieves a PSNR of 27.53 dB, 
establishing a new state-of-the-art (SOTA) 
with a model size of only one-third of its counterpart, 
and outperforming the same backbone without flow-depth complementarity by 0.19 dB. 
On ACID~\cite{liu2021infinite}, which primarily consists of outdoor natural scenes with large low-texture regions, JointSplat also demonstrates strong performance, highlighting its ability to generalize to challenging environments. 
These consistent and significant improvements demonstrate the effectiveness and generalization ability of the proposed probabilistic joint flow-depth complementary framework, offering a new 
baseline for high-fidelity 3D reconstruction from sparse views. 

\section{Related Work}
\label{Related Work}

\paragraph{Sparse Novel View Synthesis.}Sparse novel view synthesis refers to the task of generating images from arbitrary new viewpoints given only a limited number of input views~\cite{penner2017soft}. Traditional per-scene optimization methods for novel view synthesis~\cite{paliwal2024coherentgs,zheng2025flow} typically rely on scene-specific optimization procedures, which limit their generalizability. To address this limitation, recent studies have shifted focus towards feed-forward inference approaches~\cite{charatan2024pixelsplat,chen2024mvsplat,xu2024depthsplat,zhang2025transplat,tang2024hisplat}, 
which leverage geometric priors learned from large-scale multi-view datasets to reconstruct scenes in a single forward pass. For instance, 
DepthSplat~\cite{xu2024depthsplat} explores the complementarity between feed-forward sparse-view reconstruction and monocular depth estimation by fusing features from pretrained monocular depth models into the cost volume, significantly enhancing reconstruction performance while maintaining architectural simplicity. However, these feed-forward methods~\cite{charatan2024pixelsplat,chen2024mvsplat,xu2024depthsplat,zhang2025transplat,tang2024hisplat}, which rely exclusively on depth estimation, often suffer from mismatches in regions with repetitive patterns or weak textures, resulting in incomplete geometry or blurred details. To mitigate this issue, we incorporate optical flow as an additional cue and introduce a probabilistic optimization mechanism to scale the information fusion between depth and flow, thereby enabling more robust and complementary integration. 
\paragraph{Joint Depth and Flow Estimation for 3D Reconstruction.}In recent years, joint optical flow and depth estimation in 3D reconstruction has received increasing attention, particularly for its notable advantages in handling complex dynamic scenes and low-texture regions~\cite{yin2018geonet,liu2022camliflow,wan2023rpeflow}. In the context of 3D reconstruction, optical flow and depth information are tightly coupled through geometric constraints. Optical flow captures pixel-level motion, providing apparent motion cues, while depth estimation recovers the underlying scene geometry~\cite{roxas2018real}. 
Flow-NeRF~\cite{zheng2025flow} leverages bi-directional flow estimation conditioned on camera poses and incorporates a feature enhancement module to propagate canonical-space features into the world-space representation, thereby improving the quality of scene reconstruction.
CoherentGS~\cite{paliwal2024coherentgs} introduces a flow-guided consistency regularization term that constrains the projected Gaussian positions of corresponding pixels across views, thereby improving structural coherence and optimization stability. 
Despite these advances, joint flow-depth networks still face challenges in regions with large displacements and occlusions where flow estimation is unreliable~\cite{wang2020unsupervised}.
For example, CoherentGS~\cite{paliwal2024coherentgs} imposes a strict constraint that each Gaussian must lie along the ray from the camera center to its corresponding pixel, which inherently prevents the reconstruction of occluded regions.
To address these issues, we propose a probabilistic optimization mechanism that selectively down-scales the influence of unreliable optical flow, while leveraging depth-derived geometric priors. 


\begin{figure}
    \centering
    \includegraphics[width=1.0\linewidth]{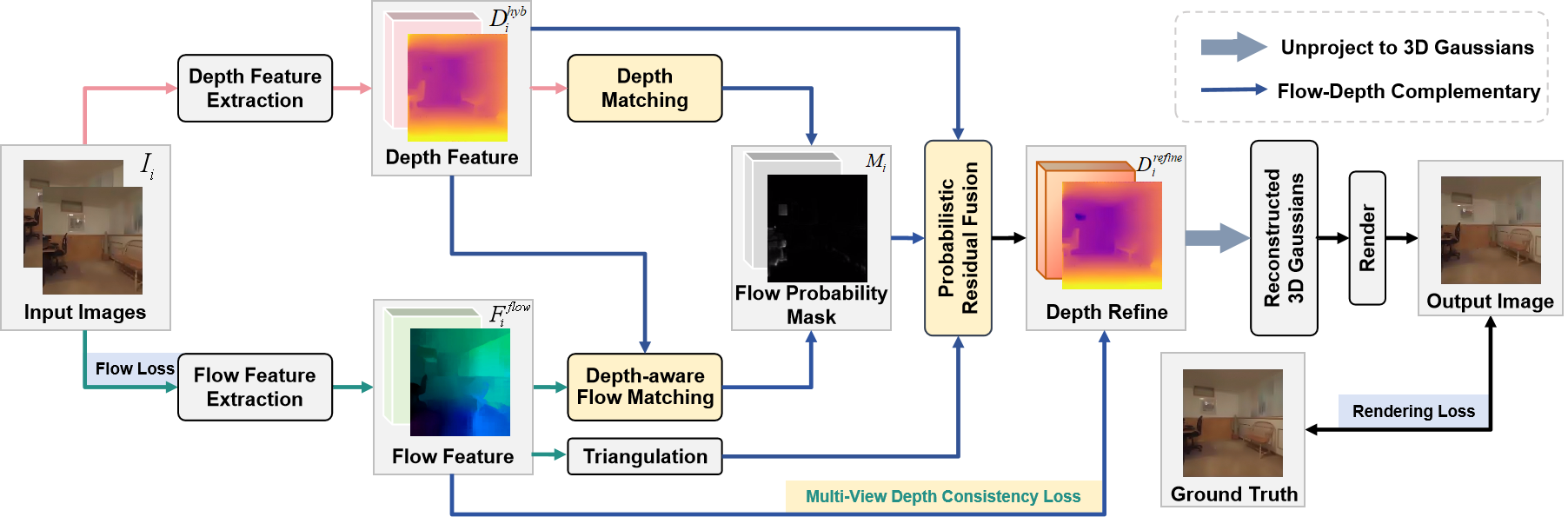}
    \caption{\textbf{The system flow of our proposed JointSplat.} 
    Given sparse-view images $I_i$ with camera intrinsics $K$ and relative poses $T_{i \to i+1}$, we extract hybrid depth $D^{\mathrm{hyb}}_i$ and optical flow features $F^{\mathrm{flow}}_i$ using pretrained models~\cite{yang2024depth,xu2023unifying,xu2022gmflow}. A probabilistic optimization mechanism is then introduced to scale the information fusion between depth and flow to obtain refined depth estimation $D^{\mathrm{ref}}_i$, which is back-projected to determine Gaussian centers for high-quality novel view synthesis.
    }
    \label{fig:framework}
    \vspace{-1em}
\end{figure}
\section{Method}
\label{Method}
Fig.~\ref{fig:framework} illustrates the system flow of our proposed JointSplat, 
which consists of two pre-trained branches: a depth branch that predicts a hybrid depth map for guiding feature matching in the flow branch and occlusion detection, and a flow branch that leverages the predicted flow and hybrid depth to estimate cross-view flow probabilities and generate flow probability masks. 
The hybrid and flow-triangulated depths are fused via the probabilistic optimization mechanism
to obtain refined depth estimates. 
The refined depth is then back-projected to determine Gaussian centers, whose attributes are predicted for high-quality novel view synthesis. 
Next, we first introduce the proposed probabilistic optimization mechanism, then present the novel view synthesis pipeline based on multi-view depth estimation, and finally, describe the proposed regularization strategies. 

Specifically, we follow the common setup in prior work~\cite{charatan2024pixelsplat,chen2024mvsplat,xu2024depthsplat}, where a set of $N$ sparse-view images $\{I_i\}_{i=1}^N$ ($I_i \in \mathbb{R}^{H \times W \times 3}$) along with their camera intrinsics $K$ and relative poses $T_{i \to i+1}$ are used to extract three types of features as inputs to the subsequent modules. Specifically, monocular features $\{F_i^{\mathrm{mono}}\}_{i=1}^N$ are obtained from the pretrained Depth Anything V2~\cite{yang2024depth} model. Multi-view cost volumes $\{C_i\}_{i=1}^N$ are constructed via GMDepth~\cite{xu2023unifying} by computing cross-view correlations. In parallel, optical flow features $\{F_i^{\mathrm{flow}}\}_{i=1}^N$ are extracted using GMFlow~\cite{xu2023unifying}, where each $F_i^{\mathrm{flow}} \in \mathbb{R}^{H \times W \times C_{\mathrm{flow}}}$, and $C_{\mathrm{flow}}$ denotes the feature channel dimension. Following the method of~\cite{xu2024depthsplat}, we employ a 2D U-Net~\cite{ronneberger2015u} and a DPT head~\cite{rombach2022high} to fuse the monocular features $F_i^{\mathrm{mono}}$ and the cost volumes $C_i$, producing the final output of our depth branch—the hybrid depth prediction ${D_i^{\mathrm{hyb}}}_{i=1}^N \in \mathbb{R}^{H \times W}$.

\subsection{Flow-Guided Geometric and Probabilistic Modeling.}
\label{3.2}

\paragraph{Depth‑Aware Flow-Matching Probability Map.}
Optical flow is fundamentally a matching problem: for every pixel in the reference image, we must locate the most similar pixel in the target image~\cite{xu2022gmflow}. 
Traditional approaches, such as~\cite{dosovitskiy2015flownet}, therefore restrict the search to a fixed local window and measure feature similarity inside that window, which keeps the computation affordable but easily fails under large viewpoint or motion changes. 
Recent works, notably GMFlow~\cite{xu2022gmflow}, reformulate optical flow as a global matching task—computing the similarity between all pairs of feature vectors to obtain dense correspondences that can handle very large displacements. 
However, performing a full matrix multiplication between two feature grids of size $H\!\times\!W$ quickly becomes prohibitive ($\mathcal{O}((HW)^2)$). 
We propose a depth‑adaptive solution that greatly reduces the search space relative to full global matching while retaining robustness to large displacements. 
Specifically, we construct a novel depth–aware flow–matching probability map $\mathbf{f_c}\in\mathbb{R}^{H\times W}$, detailed as follows.

Given a sparse‑view image set $\{I_i\}_{i=1}^{N}$, we first extract the corresponding feature maps $\mathbf{F}_i,\mathbf{F}_{i+1}\in\mathbb{R}^{H\times W\times C}$ from any adjacent image pair $(I_i,I_{i+1})$, representing views $i$ and $i{+}1$, respectively. Using the hybrid depth prediction $D_i^{\mathrm{hyb}}$ produced by fusing the two depth branches, we then normalize depth per image:
\begin{equation}
\hat{D}(u)=\frac{D_i^{\mathrm{hyb}}(u)-D_{\min}}
               {D_{\max}-D_{\min}+\varepsilon}\in[0,1],
\end{equation}
where $D_{\min}$ and $D_{\max}$ are the minimum and maximum depth values in the image, and $\varepsilon$ is the stability term, set to $10^{-6}$. 
The normalized depth then modulates the local search radius as:
\begin{equation}
r(u)=\bigl\lfloor r_{\min}+\hat{D}(u)\bigl(r_{\max}-r_{\min}\bigr)\bigr\rfloor,
\end{equation}
where $r_{\min}$ and $r_{\max}$ denote the smallest and largest radii. 
Thus, further pixels use larger windows, while closer pixels use smaller ones. This is because pixels that are further in depth typically exhibit larger apparent motion and are granted a larger search region, whereas nearby pixels are matched within a smaller neighborhood. 

For every pixel $u=(x,y)$, we therefore examine a depth–adaptive window:
\begin{equation}
\mathcal{N}(u)=\{\delta\in\mathbb{Z}^2 \mid \|\delta\|_\infty \le r(u)\},
\end{equation}
which we implement efficiently by masking a pre‑generated full window of size $(2r_{\max}+1)^2$. 

Following GMFlow~\cite{xu2022gmflow}, we compute feature similarities using dot-product operations~\cite{wang2020learning} and employ a differentiable matching layer~\cite{kendall2017end} to generate the pixel-wise matching distribution. Target‑view features are bilinearly sampled at $u+\delta$ and correlated with the reference feature by a scaled dot product~\cite{vaswani2017attention} as:
\begin{equation}
s_{u,\delta}=\langle\mathbf{F}_i(u),\mathbf{F}_{i+1}(u+\delta)\rangle/\sqrt{C},\end{equation}where $\sqrt{C}$ is a normalization factor to avoid large values. 

For every pixel location $u\!=\!(x,y)$, we compute a scaled dot‑product similarity with all valid offsets $\delta\!\in\!\mathcal{N}(u)$,
apply a softmax to obtain the matching probability $p_{u,\delta}$,
and aggregate the expected displacement to form the flow vector
$\mathbf{f}_{\mathrm{c}}(u)\!\in\!\mathbb{R}^{H\times W}$. 
The per-pixel probability $\mathbf{f_c}(u)$ is defined as the maximum of the matching probability $p_{u,\delta}$:
\begin{equation}
\mathbf{f_c}(u) = \max_{\delta \in \mathcal{N}(u)} \left\{ \frac{\exp(s_{u,\delta})}{\sum_{\delta' \in \mathcal{N}(u)} \exp(s_{u,\delta'})} \right\}, \qquad \sum_{\delta \in \mathcal{N}(u)} p_{u,\delta} = 1.
\end{equation}
which measures how unambiguously pixel $u$ can be matched across views.

\paragraph{Depth-based Occlusion Mask.}
Previous methods for evaluating per-pixel optical flow reliability often employ occlusion masks. For instance, some approaches~\cite{paliwal2024coherentgs,sundaram2010dense} use inconsistencies in the forward and backward optical flows estimated by a network to filter out unreliable information, while other methods~\cite{wang2020unsupervised} infer optical flow from depth and camera pose, and then utilize flow consistency to identify occlusion regions and generate corresponding masks. 
In contrast, our method determines occlusion by computing the pixel-level feature correlation across different views. 

Given the corresponding feature maps $\mathbf{F}_i$ and $\mathbf{F}_{i+1}$ extracted from views $i$ and $i{+}1$, the hybrid depth prediction $D_i^{\mathrm{hyb}}$ of view $i$, the intrinsic matrix $K$, and the relative pose $T_{i\!\rightarrow\! i+1}$, a per‑pixel occlusion probability is estimated in two steps. 
We first adopt depth-guided warping~\cite{xu2023unifying} to warp the feature map $\mathbf{F}_{i+1}$ back to the viewpoint of $\mathbf{F}_i$ based on the predicted depth and camera poses:
\begin{equation}
\tilde{\mathbf{F}}_{i+1}(u)=
\mathbf{F}_{i+1}\!\Bigl(
  \pi\!\bigl(
    T_{i\rightarrow i+1}\,[D_i^{\mathrm{hyb}}(u)\,K^{-1}\,\tilde u;\,1]
  \bigr)
\Bigr),
\label{eq:warp}
\end{equation}
where $\tilde u$ is the homogeneous pixel coordinate and $\pi(\cdot)$ denotes perspective projection. 
Here, by appending a constant 1, the Euclidean point is converted to homogeneous coordinates, enabling the 4×4 transform to simultaneously apply rotation and translation. 
Then, we compute a matching score for each pixel by taking the dot product between $\mathbf{F}_i(u)$ and the warped feature $\tilde{\mathbf{F}}_{i+1}(u)$, and then apply a sigmoid function after normalization:
\begin{equation}
M_{\mathbf{occ}}(u)=\mathbb{1}\!\Biggl( \sigma\!\Bigl(\frac{\langle \mathbf{F}_i(u),\,\tilde{\mathbf{F}}_{i+1}(u) \rangle}{\sqrt{C}}\Bigr)>\tau \Biggr),
\qquad \tau=0.5,
\end{equation}
where $\mathbb{1}\{\cdot\}$ is the indicator function that returns 1 if the condition is true and 0 otherwise. Here, similar to the flow matching probability map, we use the dot product and softmax (implicit in the binary thresholding) to compute the matching probability between $\mathbf{F}i(u)$ and $\tilde{\mathbf{F}}{i+1}(u)$, and pixels with low visibility are treated as occluded. The mask $M_{\mathbf{occ}}\in{0,1}^{H\times W}$ effectively suppresses unreliable correspondences caused by depth discontinuities and self‑occlusions.

\paragraph{Flow Probability Mask.}
Typically, an occlusion mask is a binary mask that indicates the valid regions in the optical flow. In prior methods~\cite{paliwal2024coherentgs,sundaram2010dense}, a value of 1 in the occlusion mask is regarded as a region with absolute optical flow confidence, while a value of 0 indicates an untrusted region.
Unlike prior methods, we propose a flow probability mask that is composed of two components: the occlusion mask that discriminates occluded regions and the flow matching probability map that estimates the reliability of the optical flow. These two components are combined via element-wise multiplication to yield the final flow probability mask:
\begin{equation}
M_{\mathbf{flow}}(u) = M_{\mathbf{occ}}(u) \cdot f_c(u).
\end{equation}
Consequently, pixels in occluded regions are assigned a value of 0, while the remaining pixels have values in the interval $(0,1]$, measuring the optical flow probability for each pixel. 

\subsection{Novel View Synthesis via Multi-View Depth Estimation}
\label{3.3}

\paragraph{Depth Completion via Flow-based Triangulation.}
Given the optical flow $F_i^{\mathrm{flow}}$ between two consecutive frames $I_i$ and $I_{i+1}$, and the intrinsic matrix $\mathbf{K}$ along with the relative camera transformation $T_{i\rightarrow i+1} = (\mathbf{R}_{i}^{i+1},\, \mathbf{t}_{i}^{i+1})$, we estimate the depth $\bar{d}_j^i$ for each pixel $\mathbf{u}_j^i = [x_j^i, y_j^i, 1]^T$ in the reference frame $i$ by solving the following least-squares triangulation problem:
\begin{equation}
\bar{d}_j^i = \arg\min_{d} \left\| \mathbf{u}_j^{i+1} \times \mathbf{K} \left( \mathbf{R}_{i}^{i+1} (d\, \mathbf{K}^{-1} \mathbf{u}_j^i) + \mathbf{t}_{i}^{i+1} \right) \right\|^2,
\label{eq:triangulation}
\end{equation}
where $\mathbf{u}_i^{t+1} = \mathbf{u}_j^i + F_i^{\mathrm{flow}}$ is the matched pixel in the target frame according to the predicted optical flow, and $\times$ denotes the vector cross product. The term $d\, \mathbf{K}^{-1} \mathbf{u}_j^i$ represents the back-projected 3D point in the camera coordinate system of frame $i$. 
This triangulation assumes a static scene and camera ego-motion, and may fail in dynamic regions~\cite{guizilini2022learning}. 
To avoid this issue, we incorporate the flow probability mask \(M_{\mathbf{flow}}(u)\) during fusion to differentiate dynamic regions from static ones and to serve as a reference for determining the fusion weights. 

To refine the estimated hybrid depth map, we predict a residual correction $\Delta(u)$ from the concatenated input consisting of the hybrid depth, flow-guided depth, and flow probability mask. This is achieved via two consecutive $3 \times 3$ convolutional layers with a ReLU activation function in between:
\begin{equation}
\Delta(u) = \mathrm{Conv}_2\left( \mathrm{ReLU}\left( \mathrm{Conv}_1\left( \left[ D_i^{\mathrm{hyb}}(u),\, D_i^{\mathrm{flow}}(u),\, M_{\mathbf{flow}}(u) \right] \right) \right) \right),
\end{equation}
where $\mathrm{Conv}_1$ and $\mathrm{Conv}_2$ denote convolutional layers, and $\mathrm{ReLU}(\cdot)$ represents the nonlinear activation. 
The final refined depth is obtained by adding the predicted residual to the hybrid estimate:
\begin{equation}
D_i^{\mathrm{refine}}(u) = D_i^{\mathrm{hyb}}(u) + \Delta(u).
\end{equation}
This residual fusion strategy enables the network to adaptively adjust the hybrid depth based on pixel-wise probabilistic optimization, thereby improving the robustness and accuracy of depth prediction, particularly in cases of insufficient local structural modeling and distortion-prone occluded regions.

\paragraph{Gaussian Parameter Prediction.}
Following~\cite{xu2024depthsplat}, for the task of 3D Gaussian splatting, we first unproject the refined depth map \(D_i^{\mathrm{refine}}(u)\) obtained via flow completion into 3D space using the camera parameters, treating the result as the Gaussian centers \(\mu_j\). The remaining Gaussian parameters are predicted by a DPT head~\cite{ranftl2021vision}, which takes as input the concatenated image, depth, and feature information, and outputs the opacity \(\alpha_j\), covariance \(\Sigma_j\), and color \(c_j\). Finally, novel views \(I_{\mathrm{render}}\) are rendered via the Gaussian splatting operation~\cite{kerbl20233d}.

\subsection{Synthetic Supervision and Real-World Self-Supervision}
\label{3.4}

\paragraph{Self-Supervised Optical Flow Estimation Loss.}
To reduce 
reliance on ground-truth optical flow annotations and improve flow estimation quality, 
we introduce a census loss $L_{\mathrm{census}}$~\cite{meister2018unflow}. 
In addition, we adopt an edge-aware smoothness loss~\cite{wang2018occlusion}, denoted as $\mathcal{L}_s$, to encourage spatial consistency of the predicted flow, and 
use both first-order and second-order derivatives to ensure the flow 
aligns with that of neighboring pixels that are most similar in appearance.

\paragraph{Multi-View Depth Consistency Loss.}
Inspired by~\cite{paliwal2024coherentgs}, we leverage optical flow to enforce the view consistency of depth values, which in turn facilitates consistent alignment of Gaussian centers across views. First, given a refined depth map $D^{\mathrm{refine}}_{i+1}$ and an optical flow $\mathbf{F}^{\mathrm{flow}}_i(u)$ representing pixel displacements from $D^{\mathrm{refine}}_i$ to $D^{\mathrm{refine}}_{i+1}$, we aim to reconstruct a warped version of $D^{\mathrm{refine}}_{i+1}$ such that it aligns with the reference view $D^{\mathrm{refine}}_i$. The warping 
is guided by the inverse flow, which ``pulls'' pixels from $D^{\mathrm{refine}}_{i+1}$ back to their corresponding locations in $D^{\mathrm{refine}}_i$:
\begin{equation}
D^{\mathrm{refine}}_i(u) \approx D^{\mathrm{refine}}_{i+1}(u + \mathbf{F}^{\mathrm{flow}}_i(u)).
\end{equation}
Then, we compute the pixel-wise reweighted depth consistency loss between the reference depth map and the backward-warped target depth map, where the per-pixel weights are determined by the probabilistic optimization mechanism as:
\begin{equation}
\mathcal{L}_{\mathrm{mvd}} = M_{\mathrm{flow}}\cdot \left| D^{\mathrm{refine}}_i - \hat{D}^{\mathrm{refine}}_i \right|,
\end{equation}
where $\hat{D}^{\mathrm{refine}}_i(u) = D^{\mathrm{refine}}_{i+1}(u + \mathbf{F}^{\mathrm{flow}}_i(u))$ is the backward-warped depth. 

Unlike prior methods~\cite{paliwal2024coherentgs,sundaram2010dense} that utilize binary forward-backward consistency masks, our flow probability mask $M_{\mathrm{flow}}(u)$ is constructed by combining an occlusion map and a continuous-valued flow probability map. Specifically, pixels in occluded regions are assigned a probability of 0, while non-occluded regions have values in the range $(0, 1]$, indicating the reliability of the estimated optical flow. The higher the probability value at a pixel, the more reliable the optical flow at that location is considered to be. Consequently, the corresponding consistency loss is amplified, enforcing stronger supervision. Conversely, unreliable or occluded regions receive lower penalties, thus avoiding the propagation of erroneous gradients. This design allows our consistency constraint to focus learning on geometrically reliable regions, thereby improving depth accuracy.

\paragraph{Rendering Loss.}
Similar to~\cite{xu2024depthsplat}, we use a combination of MSE and LPIPS~\cite{zhang2018unreasonable} to minimize the color differences between the rendered and ground truth images:
\begin{equation}
\mathcal{L}_{\mathrm{gs}} = \sum_{m=1}^M \left( \mathrm{MSE}(I^m_{\mathrm{render}}, I^m_{\mathrm{gt}}) + \lambda \cdot \mathrm{LPIPS}(I^m_{\mathrm{render}}, I^m_{\mathrm{gt}}) \right),
\label{loss:gs}
\end{equation}
where $M$ denotes the number of novel views rendered in each forward pass, and $\lambda$ is set to 0.05.

Finally, our total loss is defined as a weighted combination of several components:
\begin{equation}
\mathcal{L}_{\mathrm{total}} = 
\lambda_{s1} \cdot \mathcal{L}_s^{(1)} + 
\lambda_{s2} \cdot \mathcal{L}_s^{(2)} + 
\lambda_c \cdot L_{\mathrm{census}} + 
\lambda_g \cdot \mathcal{L}_{\mathrm{gcc}} + 
\lambda_m \cdot \mathcal{L}_{\mathrm{mvd}} + 
\mathcal{L}_{\mathrm{gs}},
\label{eq:total_loss}
\end{equation}
where $\lambda_{s1}$, $\lambda_{s2}$, $\lambda_c$, $\lambda_g$, and $\lambda_m$ are hyper-parameters. In our implementation, we set $\lambda_{s1} = \lambda_{s2} = 0.0025$ and $\lambda_c = \lambda_g = \lambda_m = 0.1$. The hyper-parameter values follow prior work to ensure consistency with established baselines and fair comparison.

\section{Experiments}
\label{Experiments}

\paragraph{Implementation Details.}
We adopt the ViT-B variant of DepthSplat's monocular branch~\cite{xu2024depthsplat} as the backbone in our method. Following the settings of DepthSplat~\cite{xu2024depthsplat}, we resize the input images to a resolution of 256×256 and optimize the model for 400K iterations using the Adam optimizer~\cite{kingma2014adam}. 
All experiments are conducted using three NVIDIA L40 GPUs with 48 GB of memory each, with a batch size of 4. The learning rate for the backbone network is set to $2 \times 10^{-4}$. We use the pretrained Depth Anything V2~\cite{yang2024depth} model as the initialization for the monocular depth branch. The GMFlow and GMDepth~\cite{xu2023unifying} models, pretrained on the FlyingThings dataset~\cite{mayer2016large}, are used as the initial weights for the optical flow branch and the multi-view depth branch, respectively. 
\paragraph{Dataset.}
The proposed JointSplat is trained and evaluated on RealEstate10K~\cite{zhou2018stereo}, a large-scale collection of YouTube videos featuring both indoor and outdoor scenes. Following previous work~\cite{xu2024depthsplat}, we use 67,477 scenes for training and 7,289 scenes for testing. To assess the cross-dataset generalization ability, we perform zero-shot evaluation on the ACID dataset~\cite{liu2021infinite}, which consists of aerial landscape videos captured by drones. 

\subsection{Main Results}
\begin{table*}[t]
  \caption{\textbf{Quantitative comparison on RealEstate10K and ACID.} Best results are in \textbf{bold}; second best are \underline{underlined}.}
  \label{tab:quantitative}
  \centering
  \begin{tabular}{l ccc ccc}
    \hline
    \multirow{2}{*}{Method} & \multicolumn{3}{ c }{RealEstate10K} & \multicolumn{3}{ c}{ACID} \\
                            & \textbf{$\mathrm{PSNR}\,\uparrow$} & \textbf{$\mathrm{SSIM}\,\uparrow$} & \textbf{$\mathrm{LPIPS}\,\downarrow$} & \textbf{$\mathrm{PSNR}\,\uparrow$} & \textbf{$\mathrm{SSIM}\,\uparrow$} & \textbf{$\mathrm{LPIPS}\,\downarrow$} \\
    \hline
    pixelNeRF~\cite{yu2021pixelnerf}  & 20.43 & 0.589 & 0.550 & 20.97 & 0.547 & 0.533 \\
    GPNR~\cite{suhail2022generalizable}       & 24.11 & 0.793 & 0.255 & 25.28 & 0.764 & 0.332 \\
    AttnRend~\cite{du2023learning}   & 24.78 & 0.820 & 0.213 & 26.88 & 0.799 & 0.218 \\
    MuRF~\cite{xu2024murf}       & 26.10 & 0.858 & 0.143 & 28.09 & 0.841 & 0.155 \\
    \hline
    PixelSplat~\cite{charatan2024pixelsplat} & 25.89 & 0.858 & 0.142 & 27.64 & 0.830 & 0.160 \\
    MVSplat~\cite{chen2024mvsplat}    & 26.39 & 0.869 & 0.128 & 28.15 & 0.841 & 0.147 \\
    TranSplat~\cite{zhang2025transplat}  & 26.69 & 0.875 & 0.125 & 28.17 & 0.842 & 0.146 \\
    HiSplat~\cite{tang2024hisplat}    & 27.21 & \underline{0.881} & 0.117 & \textbf{28.66} & \textbf{0.850} & \textbf{0.137} \\
    DepthSplat~\cite{xu2024depthsplat} & \underline{27.47} & \textbf{0.889} & \underline{0.114} & \underline{28.37} & 0.847 & \underline{0.141} \\
    \textbf{Ours}       & \textbf{27.53} & \textbf{0.889} & \textbf{0.113} & \underline{28.37} & \underline{0.848} & \underline{0.141} \\
    \hline
  \end{tabular}
  \vspace{-1em}
\end{table*}
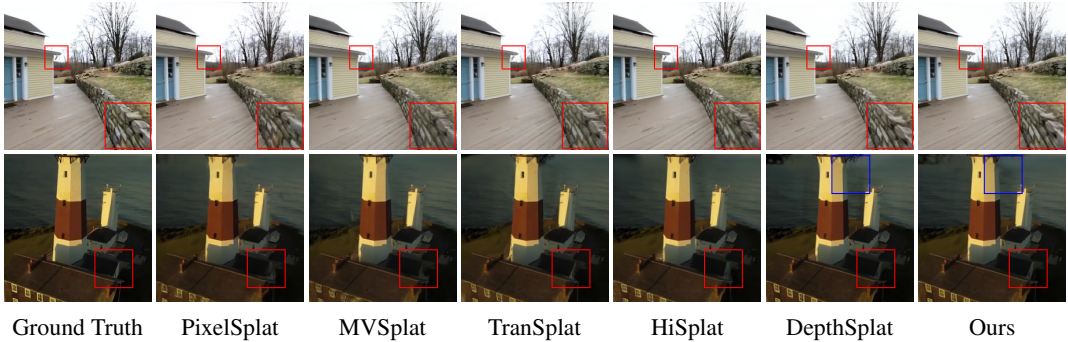
\begin{figure}[t]
    \centering
    \setlength{\tabcolsep}{1pt} 
    \renewcommand{\arraystretch}{0.8} 

    \begin{tabular}{ccccccc}
        \begin{minipage}[b]{0.14\linewidth}
            \centering
            \begin{overpic}[width=\linewidth]{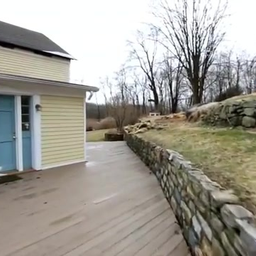}
                \put(28,55){\color{red}\framebox(15,15){}}
                \put(69,1){\color{red}\framebox(30,30){}}
            \end{overpic}
        \end{minipage}
        &
        \begin{minipage}[b]{0.14\linewidth}
            \centering
            \begin{overpic}[width=\linewidth]{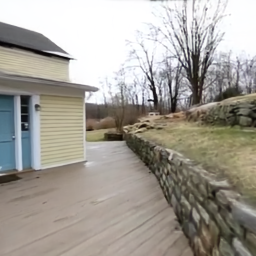}
                \put(28,55){\color{red}\framebox(15,15){}}
                \put(69,1){\color{red}\framebox(30,30){}}
            \end{overpic}
        \end{minipage}
        &
        \begin{minipage}[b]{0.14\linewidth}
            \centering
            \begin{overpic}[width=\linewidth]{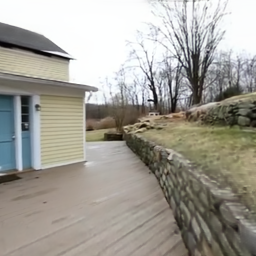}
                \put(28,55){\color{red}\framebox(15,15){}}
                \put(69,1){\color{red}\framebox(30,30){}}
            \end{overpic}
        \end{minipage}
        &
        \begin{minipage}[b]{0.14\linewidth}
            \centering
            \begin{overpic}[width=\linewidth]{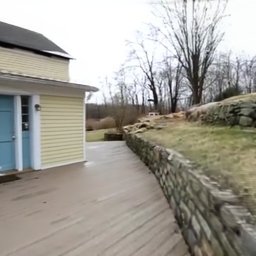}
                \put(28,55){\color{red}\framebox(15,15){}}
                \put(69,1){\color{red}\framebox(30,30){}}
            \end{overpic}
        \end{minipage}
        &
        \begin{minipage}[b]{0.14\linewidth}
            \centering
            \begin{overpic}[width=\linewidth]{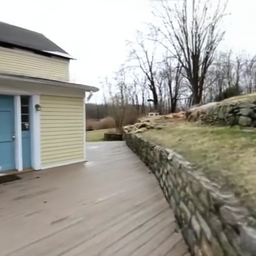}
                \put(28,55){\color{red}\framebox(15,15){}}
                \put(69,1){\color{red}\framebox(30,30){}}
            \end{overpic}
        \end{minipage}
        &
        \begin{minipage}[b]{0.14\linewidth}
            \centering
            \begin{overpic}[width=\linewidth]{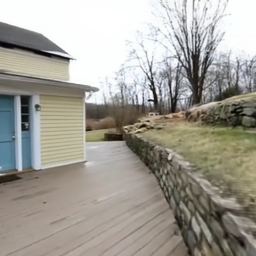}
                \put(28,55){\color{red}\framebox(15,15){}}
                \put(69,1){\color{red}\framebox(30,30){}}
            \end{overpic}
        \end{minipage}
        &
        \begin{minipage}[b]{0.14\linewidth}
            \centering
            \begin{overpic}[width=\linewidth]{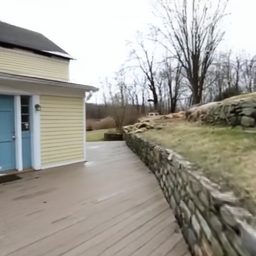}
                \put(28,55){\color{red}\framebox(15,15){}}
                \put(69,1){\color{red}\framebox(30,30){}}
            \end{overpic}
        \end{minipage}
        \\
        \begin{minipage}[b]{0.14\linewidth}
            \centering
            \begin{overpic}[width=\linewidth]{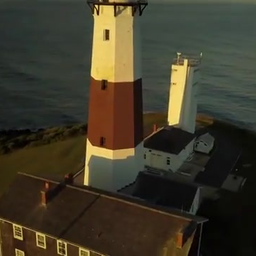}
                \put(62,10){\color{red}\framebox(25,25){}}
            \end{overpic}
        \end{minipage}
        &
        \begin{minipage}[b]{0.14\linewidth}
            \centering
            \begin{overpic}[width=\linewidth]{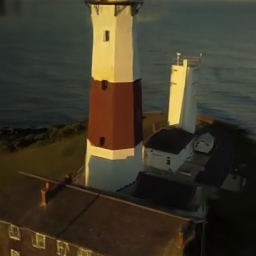}
                \put(62,10){\color{red}\framebox(25,25){}}
            \end{overpic}
        \end{minipage}
        &
        \begin{minipage}[b]{0.14\linewidth}
            \centering
            \begin{overpic}[width=\linewidth]{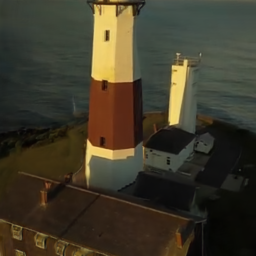}
                \put(62,10){\color{red}\framebox(25,25){}}
            \end{overpic}
        \end{minipage}
        &
        \begin{minipage}[b]{0.14\linewidth}
            \centering
            \begin{overpic}[width=\linewidth]{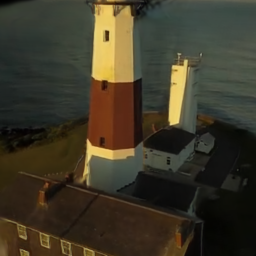}
                \put(62,10){\color{red}\framebox(25,25){}}
            \end{overpic}
        \end{minipage}
        &
        \begin{minipage}[b]{0.14\linewidth}
            \centering
            \begin{overpic}[width=\linewidth]{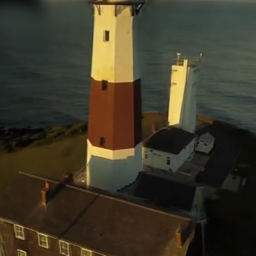}
                \put(62,10){\color{red}\framebox(25,25){}}
            \end{overpic}
        \end{minipage}
        &
        \begin{minipage}[b]{0.14\linewidth}
            \centering
            \begin{overpic}[width=\linewidth]{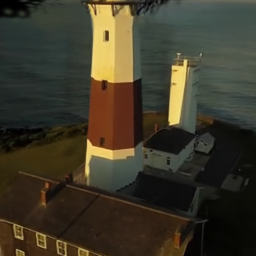}
                \put(62,10){\color{red}\framebox(25,25){}}
                \put(45,74){\color{blue}\framebox(25,25){}}
            \end{overpic}
        \end{minipage}
        &
        \begin{minipage}[b]{0.14\linewidth}
            \centering
            \begin{overpic}[width=\linewidth]{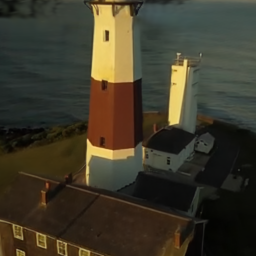}
                \put(62,10){\color{red}\framebox(25,25){}}
                \put(45,74){\color{blue}\framebox(25,25){}}
            \end{overpic}
        \end{minipage}
        \\
    \end{tabular}

    \vspace{0.3em}
    \begin{tabular}{ccccccc}
        \begin{minipage}[t]{0.14\linewidth}
            \centering
            \small Ground Truth
        \end{minipage}
        &
        \begin{minipage}[t]{0.14\linewidth}
            \centering
            \small PixelSplat
        \end{minipage}
        &
        \begin{minipage}[t]{0.14\linewidth}
            \centering
            \small MVSplat
        \end{minipage}
        &
        \begin{minipage}[t]{0.14\linewidth}
            \centering
            \small TranSplat
        \end{minipage}
        &
        \begin{minipage}[t]{0.14\linewidth}
            \centering
            \small HiSplat
        \end{minipage}
        &
        \begin{minipage}[t]{0.14\linewidth}
            \centering
            \small DepthSplat
        \end{minipage}
        &
        \begin{minipage}[t]{0.14\linewidth}
            \centering
            \small Ours
        \end{minipage}
    \end{tabular}

    \caption{\textbf{Qualitative comparison with SOTA methods.}}
    \label{fig:comparison}
    \vspace{-1em}
\end{figure}

\paragraph{Novel View Synthesis.}Tab.~\ref{tab:quantitative} presents the quantitative comparison 
of JointSplat against 
SOTA methods. Under the same training conditions, our method achieves a PSNR of 27.53 dB, an SSIM of 0.889, and a LPIPS of 0.113 on RealEstate10K~\cite{zhou2018stereo}, establishing a new SOTA performance. 

To further assess the generalization capability of JointSplat, we conduct zero-shot testing on ACID~\cite{liu2021infinite}, as shown in Tab.~\ref{tab:quantitative}. 
Unlike RealEstate10K~\cite{zhou2018stereo}, which primarily consists of indoor scenes, ACID~\cite{liu2021infinite} contains a wide range of natural landscapes characterized by large low-texture regions. Such conditions limit the effectiveness of pixel-level matching based on optical flow and result in more ambiguous geometric cues. 
Nevertheless, despite having only 125M parameters, JointSplat surpasses even the larger ViT-L version of DepthSplat~\cite{xu2024depthsplat} (354M parameters) 
with a PSNR of 28.37, SSIM of 0.848, and LPIPS of 0.141—ranking second among all evaluated methods, thereby demonstrating strong cross-domain generalization.

Qualitative visual comparisons are shown in Fig.~\ref{fig:comparison}. The first row illustrates an indoor scene from RealEstate10K~\cite{zhou2018stereo}, while the second row depicts an outdoor natural scene from ACID~\cite{liu2021infinite}. In both cases, our method produces images with accurate alignment and finer details. Notably, in the first scene, our result shows fewer artifacts than DepthSplat~\cite{xu2024depthsplat} (see the blue box). 
This improvement can be attributed to the explicit incorporation of the probabilistic optimization mechanism in our depth estimation pipeline. By fully leveraging the complementarity between optical flow and depth, and selectively enhancing reliable correspondences while suppressing noisy regions, the mechanism compensates for the limitations of previous depth-based 3D reconstruction methods in modeling local details and enables JointSplat to achieve more precise multi-view geometric alignment.
\begin{table*}[t]
  \caption{\textbf{Quantitative comparison with DepthSplat~\cite{xu2024depthsplat} on RealEstate10K~\cite{zhou2018stereo}.}}
  \label{tab:mono_ablation}
  \centering
  \begin{tabular}{l c ccc c}
    \hline
    Method & Mono branch & \textbf{$\mathrm{PSNR}\,\uparrow$} & \textbf{$\mathrm{SSIM}\,\uparrow$} & \textbf{$\mathrm{LPIPS}\,\downarrow$} & Param (M) \\
    \hline
    DepthSplat~\cite{xu2024depthsplat} & ViT-B & 27.34 & 0.887 & 0.116 & 120 \\
     & ViT-L & 27.47 & 0.889 & 0.114 & 354 \\
    \hline
    Ours                              & ViT-B & \underline{27.53} & \textbf{0.889} & \textbf{0.113} & 125 \\
                                   & ViT-L & \textbf{27.57} & \textbf{0.889} & \textbf{0.113} & 359 \\
    \hline
  \end{tabular}
  \vspace{-1em}
\end{table*}
\paragraph{Comparison with DepthSplat on RealEstate10K.}
Tab.~\ref{tab:mono_ablation} presents the quantitative comparison between our method and two versions of DepthSplat~\cite{xu2024depthsplat} on RealEstate10K~\cite{zhou2018stereo}. 
DepthSplat (ViT-B) serves as our baseline backbone, upon which we introduce optical flow features as guidance and complementary cues to further enhance performance. Specifically, with ViT-B as the backbone, our method achieves a PSNR of 27.53, SSIM of 0.889, and LPIPS of 0.113, representing improvements of +0.19 dB, +0.002, and -0.003, respectively, over DepthSplat (ViT-B).
Moreover, despite having only 125M parameters, our model surpasses even the larger ViT-L version of DepthSplat (354M parameters) across all metrics, demonstrating clear advantages in both accuracy and model efficiency. It is worth noting that although our ViT-L variant achieves slightly better results across all metrics, we primarily adopt the ViT-B configuration in our main experiments to balance performance and efficiency, given the significantly larger parameter count (359M) of ViT-L.

Fig.~\ref{fig:1} presents a qualitative comparison between our method and DepthSplat~\cite{xu2024depthsplat} for 
novel view synthesis and depth estimation. 
With the integration of probabilistic optical flow supervision, our method yields more consistent depth predictions across multiple views. This consistency is especially evident in the photo frame region (highlighted by red boxes), where View 1 and View 2 exhibit geometrically aligned depth structures—unlike the discrepancies observed in DepthSplat~\cite{xu2024depthsplat}. Furthermore, our method leverages a flow-guided triangulation mechanism that complements both monocular and multi-view depth predictions by capturing fine-grained geometric details through pixel-level motion. This design effectively compensates for the loss of local details typically seen in depth-only models, leading to cleaner and more faithful reconstructions in the synthesized novel view. As shown in the third row of Fig.~\ref{fig:1}, our rendered results not only preserve sharp structural boundaries but also exhibit fewer artifacts compared to DepthSplat~\cite{xu2024depthsplat}, further demonstrating the effectiveness of our joint flow-depth framework.
\subsection{Ablation Studies}
\paragraph{Flow Probability Mask Design.}
To assess the effectiveness of our proposed flow probability mask, we compare it against two widely used optical flow-based occlusion handling methods: the forward-backward consistency mask~\cite{sundaram2010dense} and the occlusion mask~\cite{wang2020unsupervised}. 
As shown in Tab.~\ref{tab:ablation_realestate10k_mask}, while these conventional masks perform well in dense flow settings, they tend to fail in sparse-view scenarios, where reliable forward-backward consistency is difficult to establish due to large view disparities and occlusions. Specifically, our method incorporates a flow-matching probability map to assess the reliability of optical flow at each pixel, and a depth-based occlusion mask constructed from multi-view consistency to more accurately identify regions affected by geometric inconsistencies and occlusions. In Tab.~\ref{tab:ablation_realestate10k_mask}, our full model achieves the best performance, highlighting the robustness and effectiveness of our occlusion handling strategy in enhancing geometric consistency and synthesis quality in sparse-view 3D reconstruction. The visual comparison is provided in Appendix~\ref{Occlusion}.
\begin{table*}[t]
  \caption{\textbf{The effectiveness of the key modules obtained on RealEstate10K~\cite{zhou2018stereo}.}}
  \label{tab:ablation_realestate10k_mask}
  \centering
  \begin{tabular}{lccc}
    \hline
    Method &\textbf{$\mathrm{PSNR}\,\uparrow$} & \textbf{$\mathrm{SSIM}\,\uparrow$} & \textbf{$\mathrm{LPIPS}\,\downarrow$} \\
    \hline
    \multicolumn{4}{l}{\textit{Mask Design Variants}} \\
    \quad w/ Occlusion Mask~\cite{wang2020unsupervised} & 26.71 & 0.875 & 0.124 \\
    \quad w/ Forward-Backward Consistency Mask~\cite{sundaram2010dense} & 27.09 & 0.882 & 0.119 \\
    \quad w/o Flow Matching Probability Map & 27.12 & 0.884 & 0.119 \\
    \quad w/o Depth-Based Occlusion Mask & 27.27 & 0.886 & 0.117 \\
    \hline
    \multicolumn{4}{l}{\textit{Key Module Ablations}} \\
    \quad w/o Flow Probability Mask  & 27.09 & 0.882 & 0.119 \\
    \quad w/o Depth Refine & 27.41 & 0.887 & 0.115 \\
    \quad w/o Multi-View Depth Consistency Loss & 27.44 & 0.888 & 0.115 \\
    \hline
    \textbf{Ours (Full)} & \textbf{27.53} & \textbf{0.889} & \textbf{0.113} \\
    \hline
  \end{tabular}
  \vspace{-1em}
\end{table*}

\paragraph{Effectiveness of the Key Modules.}
We further conduct ablation studies to assess the contributions of individual modules in our framework, as presented in Tab.~\ref{tab:ablation_realestate10k_mask}. Specifically, in the ``w/o Flow Probability Mask'' setting, we replace our designed flow probability mask with the forward-backward consistency mask~\cite{sundaram2010dense} to assess its impact. The results indicate that removing any of the key components—such as the flow probability mask, depth refinement, or multi-view consistency loss—leads to performance degradation, demonstrating that these modules are complementary rather than redundant. The full model achieves the best results across all metrics, confirming the effectiveness of our joint flow-depth framework in enhancing depth consistency and recovering fine-grained details.

\section{Conclusion}
\label{others}
In this work, we proposed JointSplat, a unified framework for high-fidelity sparse-view 3D reconstruction that leverages the complementarity between optical flow and depth through a novel probabilistic optimization mechanism. Instead of relying on traditional occlusion detection methods based on forward-backward consistency check, our method dynamically scales the influence of flow and depth based on pixel-wise reliability, and incorporates a novel multi-view depth-consistency loss to leverage the reliability of supervision while suppressing misleading gradients in uncertain areas. 
Comprehensive experiments on benchmark RealEstate10K~\cite{zhou2018stereo} and ACID~\cite{liu2021infinite} demonstrate that JointSplat consistently outperforms SOTA methods, showcasing strong generalizability across diverse scenes. 


{
\small
\bibliographystyle{IEEEtran}
\bibliography{IEEEexample}
}

\appendix

\section{Appendix}
\subsection{Limitations}
\label{Limitations}

Despite the effectiveness of JointSplat in enhancing reconstruction fidelity under sparse-view settings, several limitations remain. First, 
the current framework involves iterative optimization with multiple components (e.g., depth-aware flow-matching, depth completion, and occlusion reasoning), which increases computational overhead and limits real-time applicability. Second, our model is pretrained on indoor-centric datasets (e.g., RealEstate10K~\cite{zhou2018stereo}), which may reduce generalization to outdoor domains such as KITTI~\cite{geiger2013vision} without domain adaptation. Addressing these issues will be critical for scaling JointSplat to real-time or resource-constrained scenarios.

\subsection{Comparison with DepthSplat on Zero-Shot Depth Estimation}

Tab.~\ref{tab:4} shows the zero-shot performance of our method compared to DepthSplat~\cite{xu2024depthsplat} on TartanAir~\cite{wang2020tartanair}, ScanNet~\cite{dai2017scannet}, and KITTI~\cite{geiger2013vision}. 
All models are pretrained on RealEstate10K~\cite{zhou2018stereo} and evaluated directly on the target datasets without any fine-tuning. We report two standard evaluation metrics: \textbf{$\mathrm{Abs\,Rel}$} (absolute relative error), defined as $|d^* - d| / d$, and \textbf{$\delta_1$} (accuracy under threshold), which denotes the percentage of pixels where $\max(d^*/d, d/d^*) < 1.25$. Here, $d^*$ represents the predicted depth, and $d$ is the ground truth depth.
\begin{table}[!htbp]
  \caption{\textbf{Quantitative comparison with DepthSplat on depth estimation.}}
  \label{tab:4}
  \centering
  \begin{tabular}{l cc cc cc}
    \hline
    \multirow{2}{*}{Method} 
      & \multicolumn{2}{c}{\textbf{TartanAir}} 
      & \multicolumn{2}{c}{\textbf{ScanNet}} 
      & \multicolumn{2}{c}{\textbf{KITTI}} \\
      & \textbf{$\mathrm{Abs\,Rel}\,\downarrow$} & \textbf{$\delta_{1}\,\uparrow$}
      & \textbf{$\mathrm{Abs\,Rel}\,\downarrow$} & \textbf{$\delta_{1}\,\uparrow$}
      & \textbf{$\mathrm{Abs\,Rel}\,\downarrow$} & \textbf{$\delta_{1}\,\uparrow$} \\
    \hline
    DepthSplat~\cite{xu2024depthsplat} 
      & 29.53 & 53.16
      & 21.51 & 76.09
      & 56.83 & 46.26 \\
    Ours
      & 21.07 & 54.91
      & 14.61 & 74.76
      & 64.58 & 20.05 \\
    \hline
  \end{tabular}
  \vspace{-1em}
\end{table}

On TartanAir~\cite{wang2020tartanair}, a synthetic dataset featuring diverse and challenging camera motions, JointSplat achieves a significantly lower Abs Rel (21.07 vs. 29.53) and marginally improved $\delta_1$ (54.91 vs. 53.16), highlighting its ability to generalize to synthetic multi-view scenes with substantial viewpoint baselines.

\begin{figure*}[t]
  \centering
  \setlength{\tabcolsep}{2pt}   
  \renewcommand{\arraystretch}{1.0} 
  \begin{tabular}{@{}lccc@{}}
    \rotatebox{90}{\hspace*{23pt}{\small Scene 1}} &
    \begin{overpic}[width=.22\linewidth]{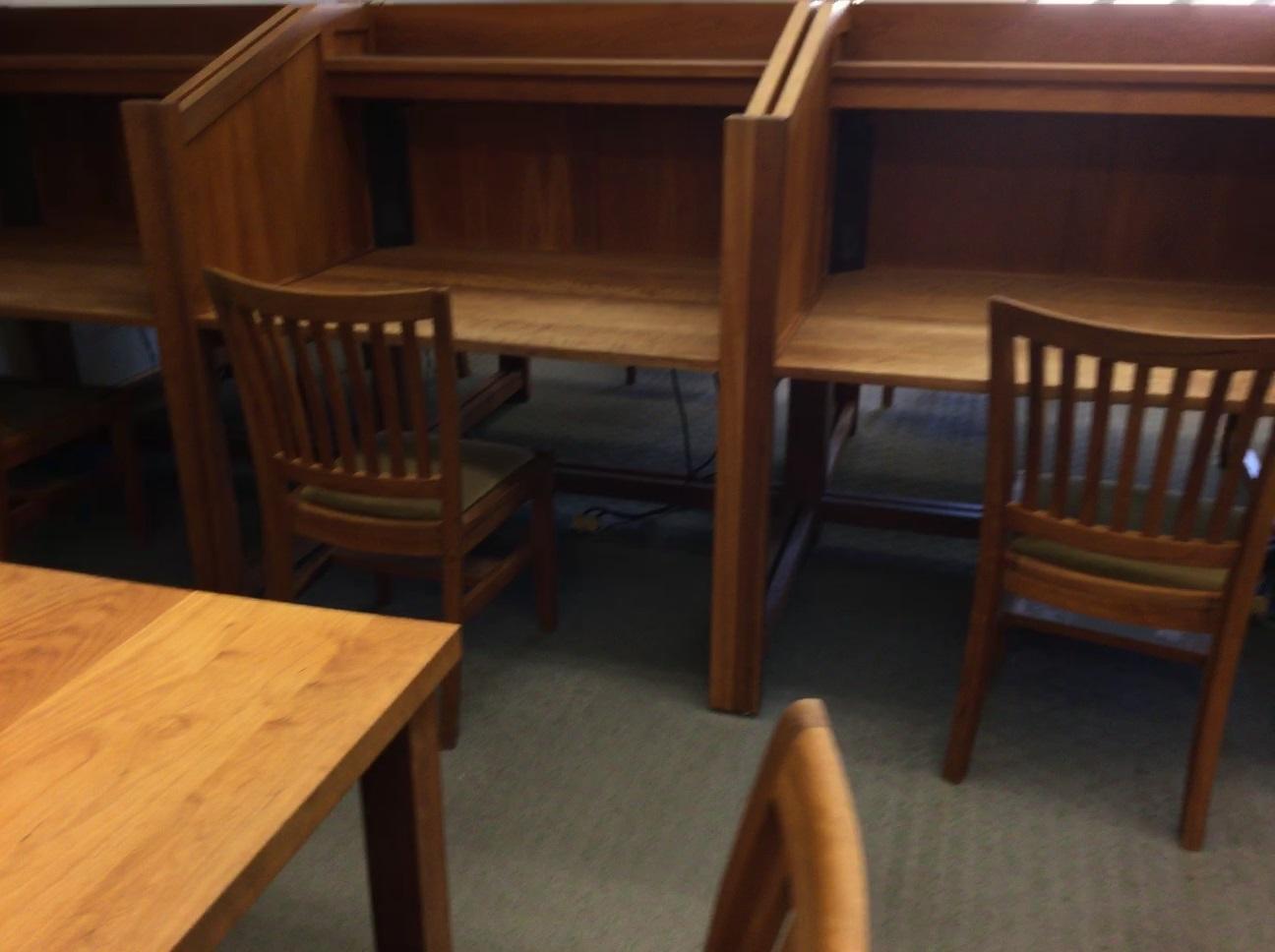}
      \put(50,25){\color{blue}\framebox(20,20){}}
    \end{overpic} &
    \begin{overpic}[width=.22\linewidth]{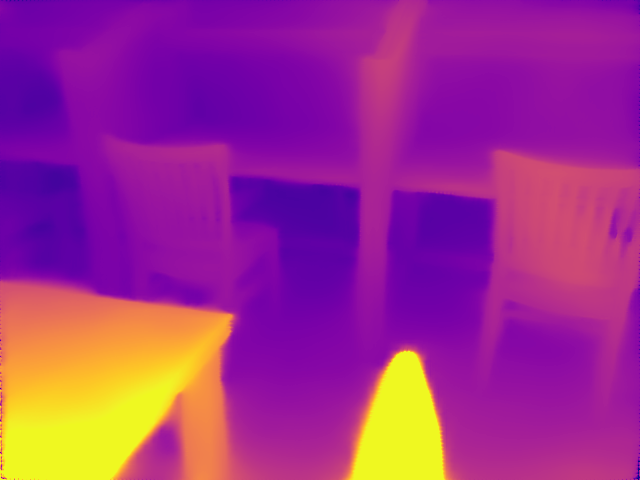}
      \put(50,25){\color{blue}\framebox(20,20){}}
    \end{overpic} &
    \begin{overpic}[width=.22\linewidth]{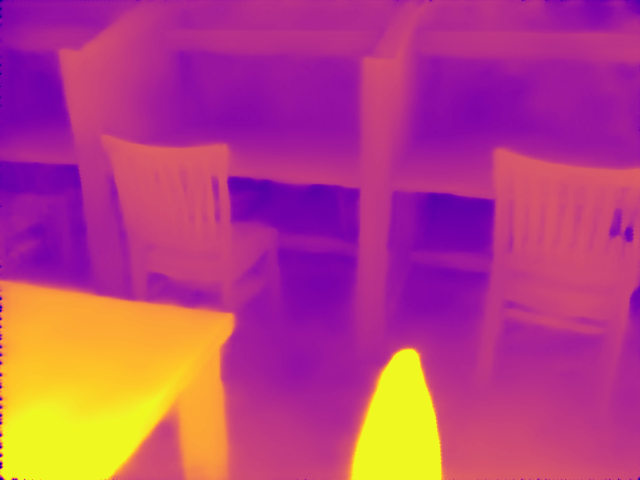}
      \put(50,25){\color{blue}\framebox(20,20){}}
    \end{overpic} \\
    \rotatebox{90}{\hspace*{23pt}{\small Scene 2}} &
    \begin{overpic}[width=.22\linewidth]{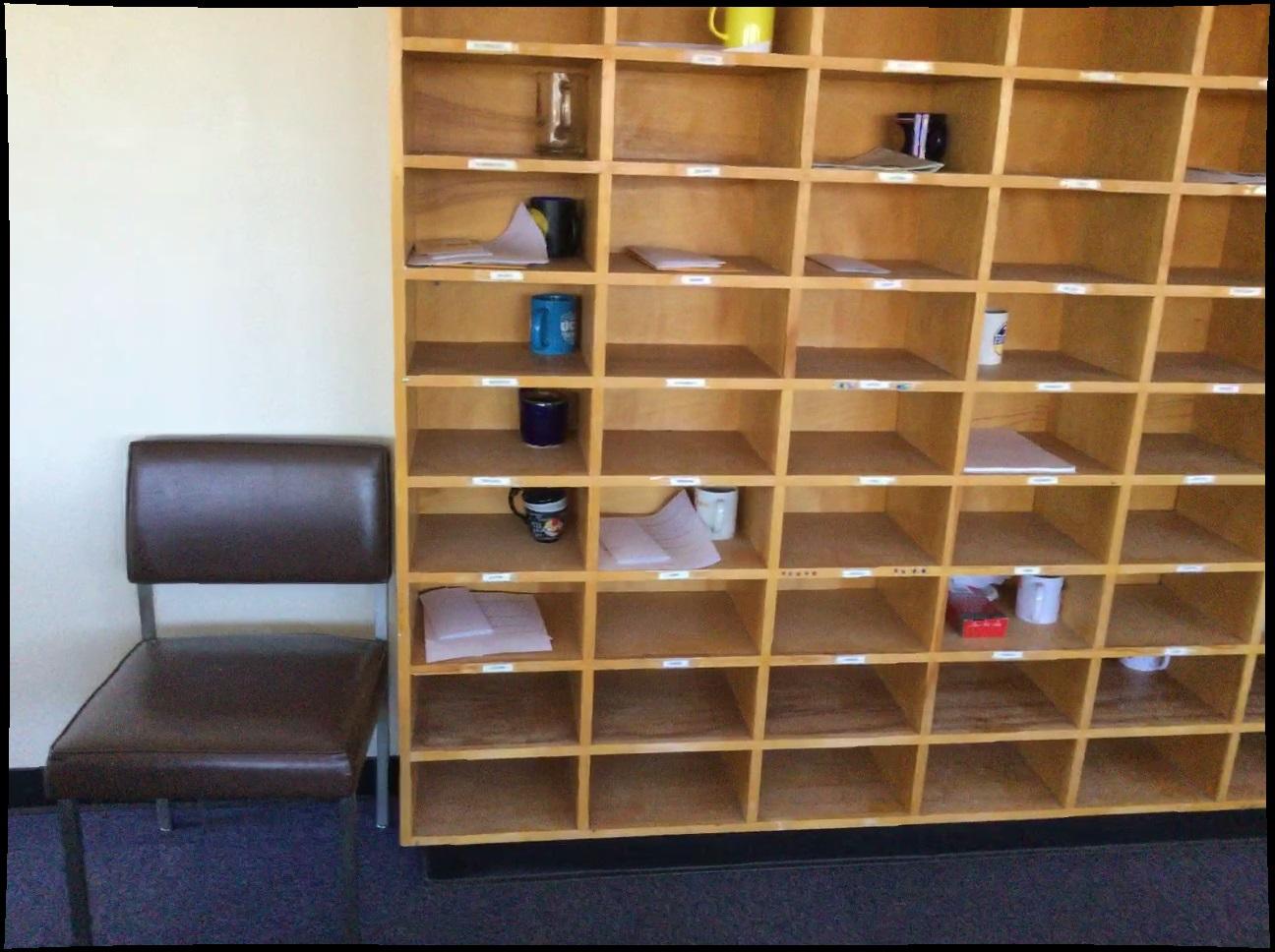}
    \put(33,28){\color{blue}\framebox(20,20){}}
    \end{overpic} &
    \begin{overpic}[width=.22\linewidth]{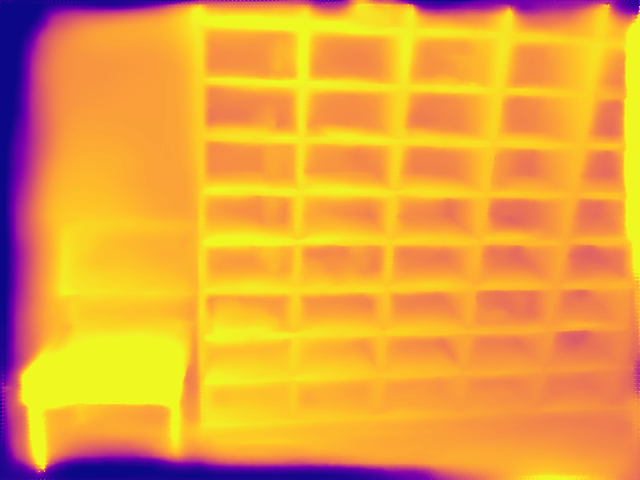}
      \put(33,28){\color{blue}\framebox(20,20){}}
    \end{overpic} &
    \begin{overpic}[width=.22\linewidth]{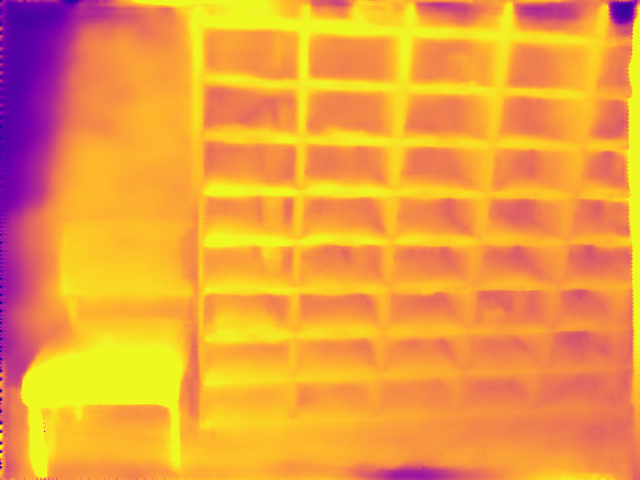}
      \put(33,28){\color{blue}\framebox(20,20){}}
    \end{overpic} \\
    {} & 
     \small Input & \small DepthSplat & \small Ours
  \end{tabular}

  \caption{\textbf{Qualitative comparison with DepthSplat~\cite{xu2024depthsplat} on depth estimation.} Blue boxes highlight regions with 
  finer structural details revealed by the proposed JointSplat. 
  Specifically, JointSplat 
  produces sharper object boundaries, better preserves geometric structures, and more accurately completes occluded or textureless areas, demonstrating improved consistency with the scene geometry.}
  \label{fig:4}
  \vspace{-1.5em}
\end{figure*}

For ScanNet~\cite{dai2017scannet}, which comprises real-world indoor environments characterized by complex geometry and frequent occlusions, our method substantially outperforms DepthSplat~\cite{xu2024depthsplat} in terms of Abs Rel (14.61 vs. 21.51). Although $\delta_1$ exhibits a slight decline (74.76 vs. 76.09), qualitative results (see Fig.~\ref{fig:4}) demonstrate superior geometric consistency, especially in the regions highlighted with blue boxes. This improvement can be attributed to the proposed probabilistic optimization mechanism, which leverages the complementarity between depth and flow. Specifically, inter-view pixel correspondence enhances stability in multi-view aggregation, while flow-based depth completion refines structural details.

In the case of KITTI~\cite{geiger2013vision}, a dataset consisting of outdoor driving scenes with long-range sparse depth and strong geometric priors, our model underperforms compared to DepthSplat~\cite{xu2024depthsplat} in both metrics (Abs Rel: 64.58 vs. 56.83, $\delta_1$: 20.05 vs. 46.26). This performance gap underscores the challenge of domain generalization from indoor-centric training data (RealEstate10K~\cite{zhou2018stereo}) to outdoor environments. The high Abs Rel suggests increased sensitivity to errors in distant depth regions; nonetheless, thanks to our proposed probabilistic joint flow-depth complementary framework, the model still retains a degree of geometric structure in local regions by leveraging pixel-wise flow matching probability to modulate supervision strength and reinforce view consistency during training.

In summary, JointSplat exhibits strong generalization across varied domains in zero-shot settings, outperforming DepthSplat~\cite{xu2024depthsplat} in synthetic and indoor scenarios, while exposing its limitations in outdoor driving contexts.

\subsection{Comparison on Occlusion Mask Design Variants}
\label{Occlusion}
To evaluate the effectiveness of occlusion handling under sparse-view settings, we compare three mask designs: the forward-backward consistency mask~\cite{sundaram2010dense}, the occlusion mask inferred from depth-flow consistency~\cite{wang2020unsupervised}, and our proposed depth-based occlusion mask, which is computed via feature correlation.
\begin{figure}[!htbp]
  \centering
  \setlength{\tabcolsep}{2pt}   
  \renewcommand{\arraystretch}{1.0} 
  \begin{tabular}{@{}lccccc@{}}
    \rotatebox{90}{\hspace*{23pt}{\small Scene 1}} &
    \includegraphics[width=.18\linewidth]{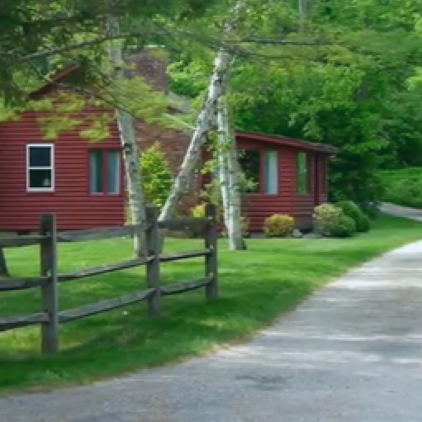} &
    \includegraphics[width=.18\linewidth]{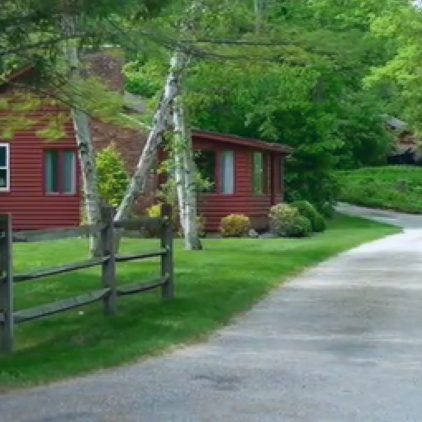} &
    \includegraphics[width=.18\linewidth]{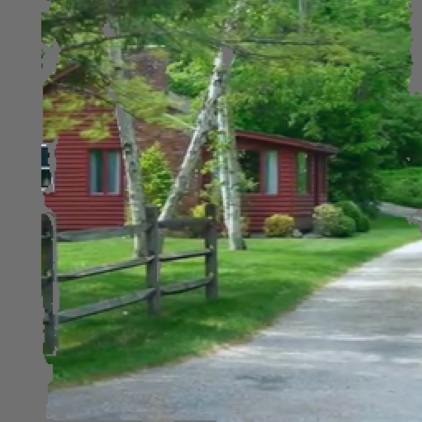} &
    \includegraphics[width=.18\linewidth]{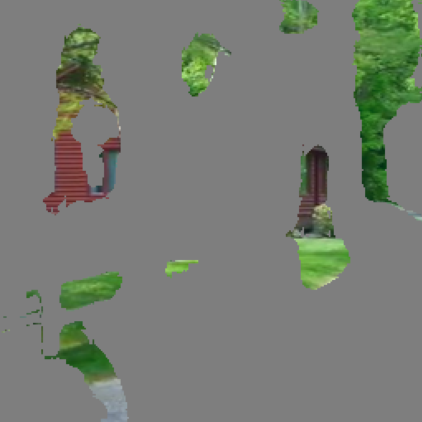} &
    \includegraphics[width=.18\linewidth]{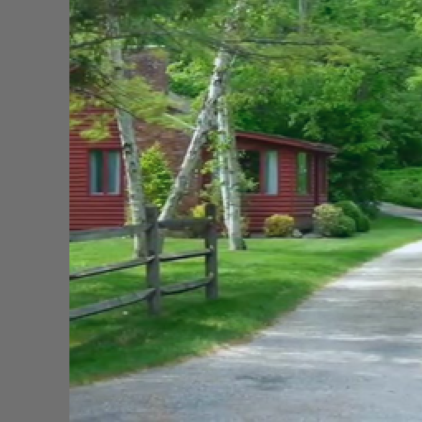} \\

    \rotatebox{90}{\hspace*{23pt}{\small Scene 2}} &
     \includegraphics[width=.18\linewidth]{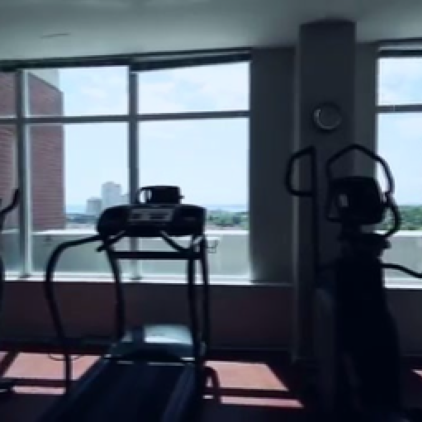} &
    \includegraphics[width=.18\linewidth]{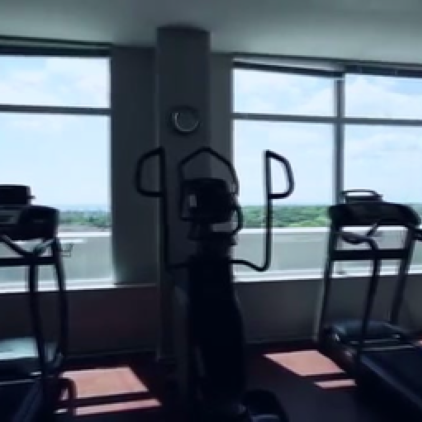} &
    \includegraphics[width=.18\linewidth]{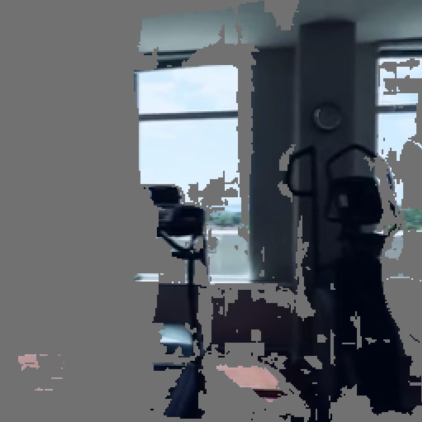} &
    \includegraphics[width=.18\linewidth]{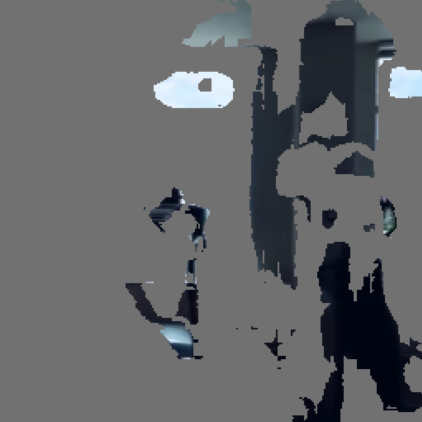} &
    \includegraphics[width=.18\linewidth]{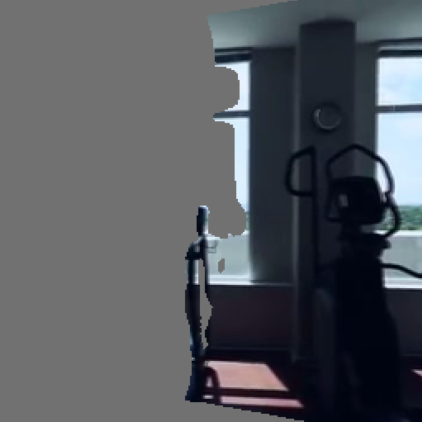} \\
     {} & 
    \small (a) View 1 & \small (b) View 2 & \small (c) FB Mask & \small (d) Occlusion Mask & \small (e) Ours
  \end{tabular}
  \caption{\textbf{Qualitative comparison of occlusion mask designs.} 
  }
  \label{fig:5methods}
  \vspace{-1.5em}
\end{figure}
As illustrated in Fig.~\ref{fig:5methods}, the occlusion mask inferred from depth-flow consistency (see Fig.~\ref{fig:5methods} (d)) tends to generate noisy and spatially incoherent predictions, often over-masking large image regions and thereby degrading reconstruction fidelity.
The forward-backward consistency mask in Fig.~\ref{fig:5methods} (c) produces sharper occlusion boundaries, but frequently exhibits over-occlusion near object contours and depth discontinuities due to unreliable flow inversions in sparse-view setups. 
In contrast, our approach estimates occlusion by measuring cross-view feature similarity at the pixel level, thereby integrating structural cues across multiple perspectives. This depth-based occlusion mask, derived from our probabilistic joint flow-depth complementary framework, suppresses low-confidence correspondences and delivers fine-grained occlusion localization around object edges. It offers more reliable occlusion awareness, which directly improves view-consistent geometry reconstruction under sparse supervision.
\end{document}